# Task Tree Retrival Algorithms for Robotic Cooking Using The Functional Object-Oriented Network

Sai Chaitanya Balli
University of south Florida
balli18@usf.edu

*Abstract*— Using the Functional Object-Oriented Network, we have implemented three search algorithms for generating the task trees for the given goal nodes. The approach, process, and the results are written in this paper.

I. INTRODUCTION

The creation of intelligent agents with the capacity to comprehend human intents and take action to solve issues in human-centered domains has received significant attention in the field of robotics research. The use of robots to aid the elderly and crippled, food delivery, and kitchen tasks are just a few examples of such sectors. However, the range of jobs and the dynamic character of the environments in which these robots will operate present the fundamental challenge in creating robots for human-centered domains. When it comes to robotic cooking, ingredients and things may come in a variety of shapes, sizes, and forms, and there are numerous states to take into consideration when following a recipe. and tables are not prescribed, although the various table text styles are provided. The formatter will need to create these components, incorporating the applicable criteria that follow. Additionally, there may be instances where a robot is unable to complete a recipe due to the lack of specific ingredients or things in its environment; this may occur when a robot is required to produce meal variations.

The Functional Object-Oriented Network (FOON), which expands on earlier work on joint object-action representation, is the knowledge representation we employ in this study. In earlier work, we showed how task planning may be accomplished by generating a FOON from video annotations. Task planning, however, is constrained to the knowledge included in a FOON because it only provides knowledge for a small number of recipe and ingredient changes, similar to previous works. For instance, there would be an issue if a robot made a salad with a specific mix of components that had never been used together before in FOON because there is no idea for that kind of salad.

We previously looked into the extension of FOON's knowledge to generalize concepts across object types. Based on this understanding, we suggest that it is possible to create new, alternative answers (as graphs) by using the current knowledge of comparable recipes. As a result, in this work, we have implemented search algorithms for generating task tree for given goal nodes, where a reference task tree is retrieved from FOON.

Our contributions and outcomes in this paper are as follows: (i) Creating the functional units for the recipe videos, (ii) Implementation of Iterative deepening search on the generated FOON graph, (iii) Implementation of Greedy Best-First Search.

II. BACKGROUND

*A. Functional Object-Oriented Network*

A FOON is a bipartite network having motion nodes and object nodes as its two different types of nodes. Edges that link objects to actions and enforce action sequencing are used to represent affordances. A core component known as a functional unit, which has input object nodes, output object nodes, and a motion node, depicts actions in FOON by specifying the state change of objects before and after execution. It is comparable to a planning operator in PDDL, where input and output nodes represent preconditions and effects, respectively.

FOONs are typically produced by annotating videos of demonstrations. A subgraph is a FOON that represents a single activity; it consists of functional units that describe the conditions of the objects before and after each action, as well as the objects that are being handled. Currently, the annotation method is manual, however earlier research looked into how graphs may be annotated semi-automatically. What we refer to as a universal FOON can be created by combining two or more subparagraphs. Once it has been combined with a variety of knowledge sources, a universal FOON can include variations of recipes. This merging technique is just a union operation performed on all functional units from each subgraph that we desire to combine; as a consequence, duplicate functional units are removed from the merged network. FOON now consists of 140 subgraph annotations of videos from YouTube, Activity-Net, and EPIC KITCHENS, all of which are accessible via our website.

*B. Basics of a FOON*

A FOON, as previously indicated, contains two sorts of nodes. This type of graph is known more technically as a bipartite network. Object nodes (denoted as NO) in a FOON are those that are manipulated in the environment or are utilized to manipulate other object nodes. In general, we only pay attention to objects that are actively employed or acted upon in a given activity. Motion nodes (denoted as NM) are another form of node that describes the manipulation of these objects. These motions can include picking and placing, pouring, chopping, and stirring. Objects can also include other objects within them; thus, we can identify objects based on their ingredient content; bowl, cup, and box objects are examples of such containers. As in cooking, motion nodes are only recognized by a motion type from a collection of pre-defined movements or actions.

Identify applicable funding agency here. If none, delete this text box.

A FOON has connections between object nodes and motion nodes as well as between object nodes and motion nodes. The only time objects can be connected to objects is when we convert our bipartite network to a one-mode projected graph for network analysis, as we did before in. Edges in our graph are drawn from one node to another in the order of a sequence that leads to the occurrence of a specific object-state outcome.

*C. The Functional Unit*

A FOON is made up of individual, essential learning units, which we call functional units. A functional unit represents a single, atomic action, and a subgraph is a collection of functional units that define an activity that involves two or more stages. A subgraph is generated to depict the complete action in a video where the demonstrator is creating macaroni and cheese, for example. This subgraph could be made up of numerous units that describe actions like boiling water in a pot, stirring macaroni pasta in a pot with a spoon, and pouring the cooked pasta into a pan.

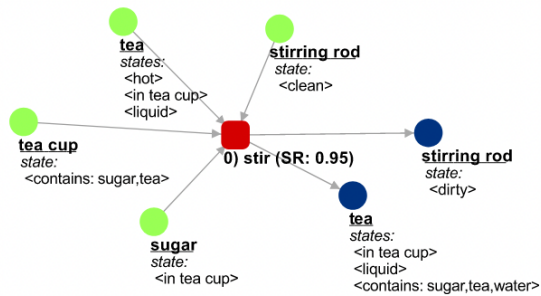

A functional unit is made up of three parts: input object nodes, output object nodes, and an intermediary motion node that defines the cause of the state transition in the objects. A motion does not guarantee that all input object states will change, hence an item may remain in the same state in successive units. As a result of the possibility of loops, a FOON can be more precisely defined as a directed acyclic graph.

## III. CREATING A FOON

A FOON is built using informational sources of knowledge, such as human activity demonstrations or observations taken from instructional movies. However, automatic information extraction from such sources is extremely challenging due to the difficulties in distinguishing the objects being used, the states they are in, and the motion that is occurring. In the absence of such a system, we annotate movies manually for the time being; volunteers were charged with the selection and annotation of cookery videos. For reference, we also take note of the timestamps at which activities occur in source videos.

*1) Gathering and Combining Knowledge*

The knowledge represented in a FOON is derived from a collection of YouTube video sources. For each source video, a subgraph will be formed in which functional units are directly constructed through hand annotation. The annotation method entails simply documenting activities in movies, specifically the time they occur, the objects being manipulated, changes in their states (if any), and the sort of motion occurring. We can then merge the knowledge from these distinct subgraphs into a single, bigger FOON using a merging technique. In theory, the merging process is extremely simple: we conduct a union operation on all functional units while deleting any duplicates. In this context, a duplicate indicates that two units contain identical input, object, and motion nodes down to the tiniest detail. We parse these files before merging to ensure that the labels are compatible with the object and motion indices preserved as references. We suggest readers to for more information on the algorithm, including pseudocode.

*2) A Universal FOON*

A universal FOON is defined as a merged set of two or more subgraphs from different information sources. Because a universal FOON is made up of knowledge from several sources, it can be used as a knowledge base by a robot to solve problems utilizing object-motion affordances. At the time of writing, our universal FOON is comprised of 65 YouTube source videos covering a variety of recipes. We give examples of the FOON graphs mentioned in this research, as well as all video subgraph files, for interested readers to download from our website.

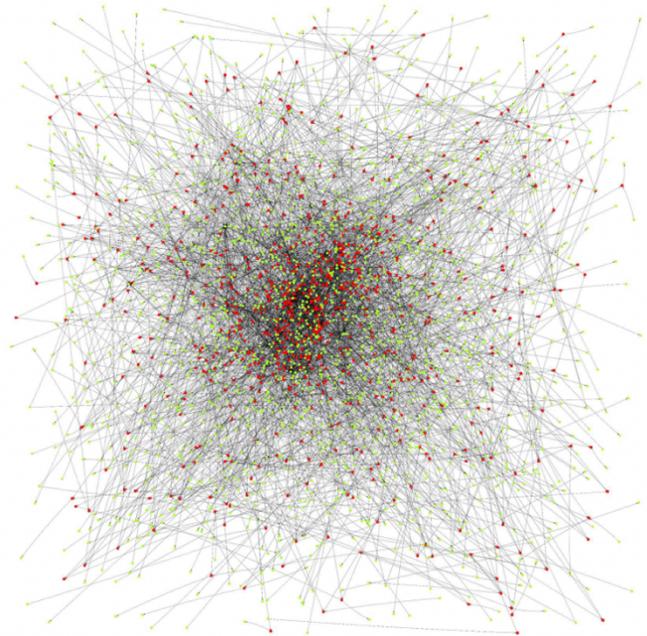

*3) Knowledge Retrieval*

A universal FOON will provide a robot with knowledge that it can use to solve manipulation tasks given a target goal. Given specific limits, a human user may instruct a robot to create a meal. The goal of knowledge retrieval is to locate a task tree: a series of functional unit-based processes that, when completed, achieve a goal. A task tree is just a collection of functional units that are likely to be linked together and that, when executed in succession, play out the execution of steps that solve a manipulation goal. This goal can be any object node in FOON, whether it is a final product or an item in an intermediate state.

Identify applicable funding agency here. If none, delete this text box.

The retrieval technique for a task tree sequence is based on the concepts of fundamental graph searching algorithms; when searching, we investigate depth-wise per functional unit, but breadth-wise across items within each unit. To solve such challenges, the robot must be knowledgeable about its domain, precisely what utensils or ingredients are in its immediate surroundings, so that the system can determine whether or not a solution exists in that case. This search yields either a task tree sequence (where a target node is deemed solvable and we have a functional unit sequence that generates the goal) or no tree owing to time constraints or a non-existent solution.

As a heuristic for locating the best task tree, we use the number of units (or steps) in our search. There may be numerous units that make up an object (for example, different trees with/without the same step size), but the search method only considers the first unit that can be executed completely (or specifically, where all objects required are available as input to that unit). We can settle ties in functional units based on job complexity instead of utilizing a step-based heuristic to find a tree. A robot may be impossible to perform a given move in some cases due to constraints in its configuration space or architecture. However, we can compensate for this by performing a simpler adjustment that yields the same results. As we continue to expand FOON, we will need to make adjustments to account for other limits, such as producing meals without

## IV. METHODOLOGY

### A. Iterative Deepening Search

Iterative Deepening Search (IDS) is an iterative graph searching approach that consumes substantially less memory in each iteration while benefiting from the completeness of the Breadth-First Search (BFS) strategy (similar to Depth-First Search). IDS accomplishes the needed completeness by imposing a depth limit on DFS, which reduces the danger of becoming stuck in an infinite or very long branch. It traverses each node's branch from left to right until it reaches the appropriate depth. After that, IDS returns to the root node and explores a separate branch that is comparable to DFS.

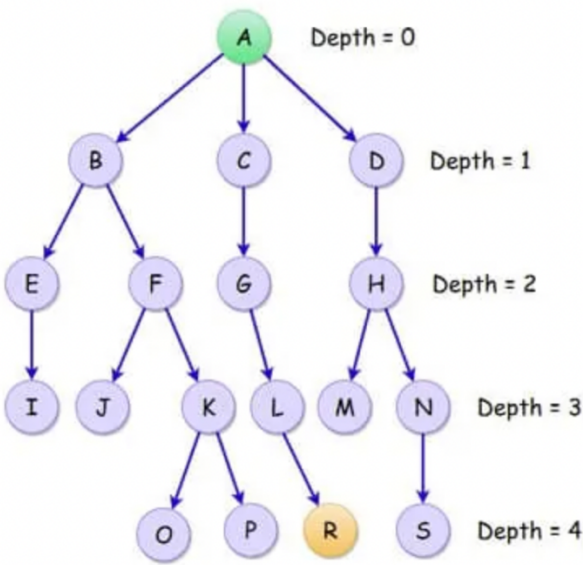

The tree can be visited as: A B E F C G D H
DEPTH = {0, 1, 2, 3, 4}

| DEPTH LIMITS | IDDFS |
|---|---|
| 0 | A |
| 1 | A B C D |
| 2 | A B E F C G D H |
| 3 | A B E I F J K C G L D H M N |
| 4 | A B E I F J K O P C G L R D H M N S |

*1) Time & space complexity:* Assume we have a tree in which each node has b children. This will be our branching factor, and d will be the tree's depth. Nodes on the lowest level, $d$, will be extended exactly once, whereas nodes on levels $d-1$ will be expanded twice. Our tree's root node will be extended $d+1$ times. If we combine all of these terms, we get:

$$(d)b+(d-1)b^2+...+(3)b^{d-2}+(2)b^{d-1}+b$$

Summation of time complexity will be: $O(b^d)$

The space complexity is: $O(bd)$, In this case, we suppose b is constant and that all children are formed at each depth of the tree and saved in a stack during DFS.

*2) Performance analysis:* It may appear that IDS has a significant overhead in the form of continuously running over the same nodes, but this is not the case. This is due to the fact that the algorithm only visits the bottom levels of a tree once or twice. Because upper-level nodes do not constitute the majority of nodes in a tree, the cost is maintained to a minimal minimum.

*3) Implementation:* we need to explore all possible paths to find the optimal solution. for makeing it simple, we just took the first path that we find. we kept increasing the depth until we find the solution. The task tree is considered a solution if the leaf nodes are available in the kitchen

```
1  IDS(root, goal_node, depthLmt){
2      for d = 0 to depthLmt
3          if (DepthFirstSearch(root, goal_node, d))
4              return true
5      return false
6  }
7
8  DepthFirstSearch(root, d){
9      if root == goal_node
10         return true
11     if d == 0
12         return false
13     for child in root.children
14         if (DepthFirstSearch(child, goal_node, d - 1))
15             return true
16     return false
17 }
```

*B. Greedy Best-First Search:*

For huge search spaces, the informed search algorithm is more useful. Because informed search algorithm employs heuristics, it is also known as Heuristic search.

**Heuristics function:** Heuristic is a function in Informed Search that finds the most promising path. It takes the agent's current state as input and calculates how close the agent is to the goal. The heuristic method, on the other hand, may not always provide the greatest solution, but it will always discover a good solution in a fair amount of time. The heuristic function calculates how close a state is to reaching the goal. It is denoted by $h(n)$, and it computes the cost of an optimal path between two states.

**Heuristics 1:** Here we are considering the motion rates for the selecting the Input nodes as the heuristic function. The basic pseudocode follows:

---

**Input:** Given Goal node G and ingredients I

**T ← A list of functional units in Task tree.**

**Q ← A queue for items to search**

**Kingd ← List of items available in kitchen.**

**Q**.push(**G**)

While **Q** is not empty **do:**

    **N** ←**Q**.dequeue()

    If **N** not in **Kingd** then:

        **C** ←Find all functional units that create **C**

        **max** = -1

        for each **candidate** in **C** do:

            if **candidate.successRate** > **max** then:

                **max** =**candidate.successRate**

                **CMax**= **candidate**

        **End for**

        **T**.append(**CMax**)

        for each **input** in **Cmax** do:

            if **n** is not visted then:

                **Q**.enque(**n**)

                **Make n visitied**

            **End if**

        **End for**

    **End if**

**End while**

**T.reverse()**

**Output: T**

---

**Heuristics 2:** This algorithm is similar to the second in that the number of input nodes and their components are considered while selecting the candidate unit. A functional unit with the fewest input nodes will be picked as a candidate unit at each level.

## DISCUSSION

An iterative deepening search explores the FOON by performing DFS and BFS at the chosen depth bound. The depth level will continue to rise until a solution is found. If the answer emerges at a deeper level, this approach requires more time to assemble the task tree. This will add to the temporal complexity by traversing all previously visited nodes for each depth-bound increment. Because they follow BFS, heuristics 1 and 2 easily locate the answer at higher levels, but each complexity increases if the solution occurs at deeper layers.

| Goal Nodes | IDS | Heuristics 1 | Heuristic 2 |
|---|---|---|---|
| Greek Salad | 31 | 32 | 28 |
| Ice | 1 | 1 | 1 |
| Macaroni | 7 | 7 | 8 |
| Sweet potato | 3 | 3 | 3 |
| Whipped Cream | 10 | 10 | 15 |

The task trees for all three methods could have the same or different numbers of functional units. All task trees have the same number of functional units for the target nodes ice and sweet potato.

## REFERENCES

[1] S. Ren and Y. Sun, "Human-object-object-interaction affordance", Proc. Workshop Robot Vis., pp. 1-6, 2013.

[2]. D. Paulius, K. S. P. Dong and Y. Sun, "Task planning with a weighted functional object-oriented network", Proc. IEEE Int. Conf. Robot. Automat., pp. 3904-3910, 2021.

[3]. D. Paulius, Y. Huang, R. Milton, W. D. Buchanan, J. Sam and Y. Sun, "Functional object-oriented network for manipulation learning", Proc. IEEE/RSJ Int. Conf. Intell. Robots Syst., pp. 2655 2662, 2016.

[4]. A. B. Jelodar and Y. Sun, "Joint object and state recognition using language knowledge", Proc. IEEE Int. Conf. Image Process., pp.

[5]. J. Marin et al., "RecipelM: A dataset for learning cross-modal embeddings for cooking recipes and food images", IEEE Trans. Pattern Anal. Mach. Intell, vol. 43, no. 1, pp. 187-203, Jan. 2021.

[6]. D. Paulius, A. B. Jelodar and Y. Sun, "Functional object-oriented network: Construction & expansion", Proc. IEEE Int. Conf. Robot. Automat., pp. 5935-5941, 2018.